\documentclass[10pt,twocolumn,letterpaper]{article}

\usepackage{cvpr}
\usepackage{times}
\usepackage{epsfig}
\usepackage{graphicx}
\usepackage{amsmath}
\usepackage{amssymb}
\usepackage{mathtools}

\usepackage[breaklinks=true,bookmarks=false]{hyperref}

\cvprfinalcopy 
\def\cvprPaperID{2497} 

\ifcvprfinal\fi
\begin{document}

\title{Spatially-Varying  Blur Detection Based on Multiscale Fused and Sorted Transform Coefficients of Gradient Magnitudes}

\author{S. Alireza Golestaneh \\
Arizona State University\\
{\tt\small sgolest1@asu.edu}
\and
Lina J. Karam\\
Arizona State University\\
{\tt\small karam@asu.edu}
}

\maketitle

\begin{abstract}
The detection of spatially-varying blur  without having any information about the blur type is a   challenging task.
In this paper, we propose a novel effective approach to address  this  blur detection problem from a single image without requiring any knowledge about the blur type, level, or camera settings.
Our approach computes blur detection maps based on a novel 
High-frequency multiscale Fusion and Sort Transform (HiFST) of gradient magnitudes. 
The evaluations of the proposed approach on a diverse set of blurry images with different blur types, levels, and contents 
demonstrate that the proposed algorithm performs favorably against the state-of-the-art methods qualitatively and quantitatively.
\end{abstract}

\section{Introduction}
Many images contain blurred regions. 
 Blur can be caused by different factors such as defocus, camera/object motion, or camera shake.
While it is common    in photography to deliberately use a shallow focus technique to give prominence to foreground objects based on defocus,  unintentional blur due to degradation factors can decrease the image quality. 
Blur detection plays an important role in many computer vision and computer graphics applications including but not limited to image segmentation, depth estimation,  image deblurring and refocusing, and background blur magnification.

\begin{figure}[!]
\begin{center}
   \includegraphics[width=1 \linewidth]{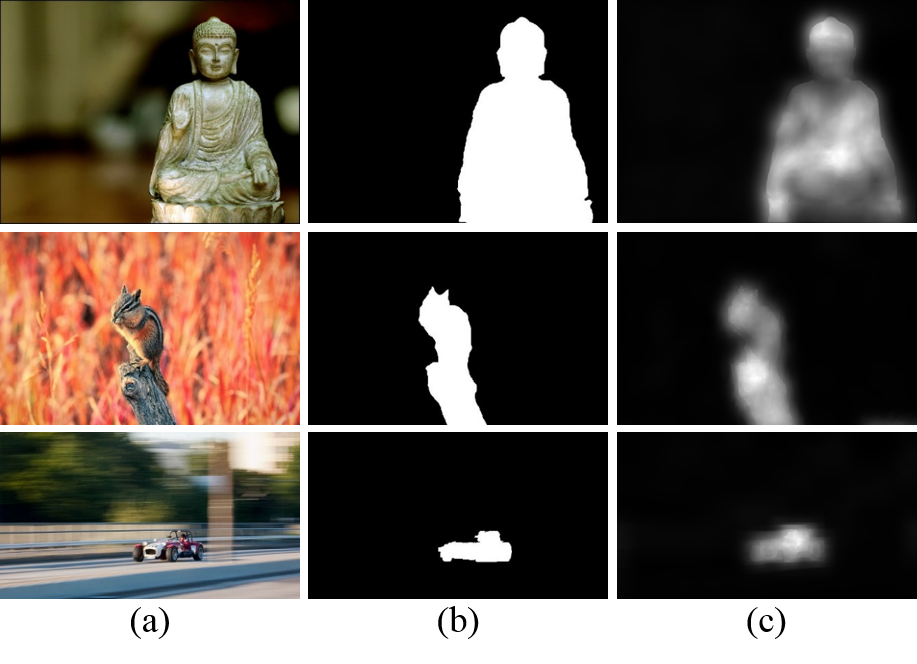}
\end{center}
   \caption{Example results of our proposed blur  detection method. 
   (a) Input images. (b) Ground-truth binary blur maps, with white corresponding to 
sharp and black corresponding to blurred region. (c) Grayscale blur detection maps generated by our proposed method with higher intensities corresponding to sharper regions.}
   \label{Pic1}
\end{figure}

In recent years, a variety of  methods have been proposed to address the issue of deblurring   by estimating blur kernels and performing a deconvolution  \cite{chakrabarti2012depth, couzinie2013learning,dai2009removing,  huimage,joshi2008psf,laicomparative,levin2009understanding,  panblind,pansoft, xu2010two,zhang2013blur,zhang2016spatially}. 
In this work, we do not aim to do kernel estimation and deconvolution. 
Instead, the objective of this work is to propose
an effective blur detection method from a single   image without having any information about the blur type, level, or the camera settings.
Figure \ref{Pic1} shows   sample results of our proposed method.

Despite the   success of existing spatially-varying blur detection methods, 
there are only   few methods  focusing on spatially-varying blur detection      regardless of the blur type \cite{chakrabarti2010analyzing,liu2008image,shi2014discriminative,
 shi2015just,su2011blurred,tang2016spectral}, and the rest  perform well only on defocus blur or motion blur.
 Moreover, the performance of most of the existing methods degrades drastically when taking into account the effects of camera noise and distortion.
Therefore, noise-free and artifact-free assumptions could be unsuitable when dealing with  real-word images.

The contributions of this work are summarized as follows. 
We propose a robust spatially-varying  blur detection method from a single image based on a novel high-frequency multiscale fusion and sort transform   (\textit{HiFST}) of gradient magnitudes to determine the level of blur at each location in an image.  
We evaluate our proposed algorithm on both defocus and motion blur types to demonstrate the effectiveness of our method.
We also test the robustness of our method by adding different levels of noise as well as different types and levels of distortions to the input image.
We compare our method with state-of-the-art algorithms using their provided implementations and demonstrate  that our proposed method
outperforms existing state-of-the-art  methods quantitatively and qualitatively.
Finally, we provide a few applications of our method including camera focus points   estimation, blur magnification, depth of field estimation, depth from focus, and deblurring.

\subsection{Related work}
Blur detection methods can be divided into two categories:
1) methods that make use of   multiple images \cite{favaro2005geometric,favaro2008shape,pentland1987new,wang2001unsupervised,zhou2010depth, zhou2009coded}, 
and 2) methods that require only a single image \cite{bae2007defocus, chakrabarti2010analyzing,chen2016fast, elder1998local,gast2016parametric,
kalalembang2009dct, liu2008image,marichal1999blur,
pentland1987new,shi2014discriminative, shi2015just,su2011blurred, tai2009single, tang2013defocus,tang2016spectral,yi2016lbp,
zhu2016efficient, zhu2013estimating,
zhuo2011defocus}.
In the first category, a set of images of the same scene are captured using multiple focus settings. 
Then the blur map is estimated during an implicit or explicit  process.
Different factors such as   occlusion   and requiring the scene to be static cause the application of these methods to be limited in practice.
In recent years, several methods have been proposed to recover a blur map from a single image without having any information about the camera settings.

In general,  blur detection algorithms from a single image can be divided into gradient-based, intensity-based and transform-based algorithms.
In \cite{chakrabarti2010analyzing},  Chakrabarti \etal propose  a sub-band decomposition based approach.
They estimate the likelihood function of a given candidate point spread function (PSF) based on local frequency component analysis.
Liu \etal\cite{liu2008image} propose a method which employs  features such as  image color, gradient, and spectrum information to  classify  blurred images. 
Shi \etal\cite{shi2014discriminative} propose a method based on    different features such as gradient histogram span, kurtosis, and data-driven local filters to differentiate between blurred and unblurred image regions. 
Shi \etal\cite{shi2015just}  propose a method  based on utilizing a sparse representation of image patches using a learned dictionary for the detection of slight perceivable blur.
In \cite{su2011blurred}, Su \etal propose a method based on examining singular value information  to measure   blurriness. 
The blur type  (motion blur or defocus blur) is then determined based on certain alpha channel constraints.
In \cite{tang2016spectral}, Tang \etal employ the image spectrum residual \cite{hou2007saliency}, 
and then they use an iterative updating mechanism  to
refine the blur map from coarse to fine by exploiting the intrinsic
relevance of similar neighboring image regions.

 In \cite{elder1998local}, Elder and Zucker propose a method that makes use of the  first- and second-order gradient information for local blur estimation.
Bae and Durand \cite{bae2007defocus} estimate the size of the blur kernel at edges, building on the method by \cite{elder1998local}, and then propagate this defocus measure over the image with a non-homogeneous optimization.
In their propagation, they assume that blurriness is smooth where intensity and color are similar.
Tai and Brown \cite{tai2009single} propose a method for estimating a defocus blur map based on the relationship between  the image contrast and  the image gradient in a local image region, namely local contrast prior. 
They use the local contrast prior to measure the defocus at each pixel and then apply Markov Random Field propagation to refine the defocus map.
In \cite{tang2013defocus},  Tang \etal use the relationship between the amount of spatially-varying defocus blur and spectrum contrast at edge
locations to estimate the blur amount at the edge locations. 
Then a defocus map  is obtained by
propagating the blur amount at edge locations over the  image using a non-homogeneous optimization procedure.
Yi and Eramian \cite{yi2016lbp} propose a local binary patterns (LBP) based method   for defocus blur segmentation  by using the distribution of uniform LBP patterns in blurred and unblurred image regions. 
Zhang and Hirakawa \cite{zhang2013blur} propose a method for estimating a defocus blur map from a single image via local frequency component analysis similar to \cite{chakrabarti2010analyzing}; they also incorporate   smoothness and color edge information into consideration to  
generate a  blur map indicating the amount of blur at each pixel. 
 Zhuo and Sim \cite{zhuo2011defocus}  compute the defocus blur from the ratio between the gradients of input and re-blurred images. 
Then they propagate the blur amount at edge locations to the entire image via matting interpolation to obtain the full defocus map.

In comparison to these methods, we introduce a new  method to estimate spatially-varying blur from a single image.
Our work is based on a multiscale transform decomposition followed by the fusion and sorting of the   high-frequency coefficients of gradient magnitudes.
The proposed method is not limited by the type of blur and does not require information about the blur type, level, or camera settings.
Experimental results demonstrate the effectiveness and robustness of our method in providing a reliable  blur detection map for different types and levels of blur.

\begin{figure}[t]
\begin{center}
   \includegraphics[width=1 \linewidth]{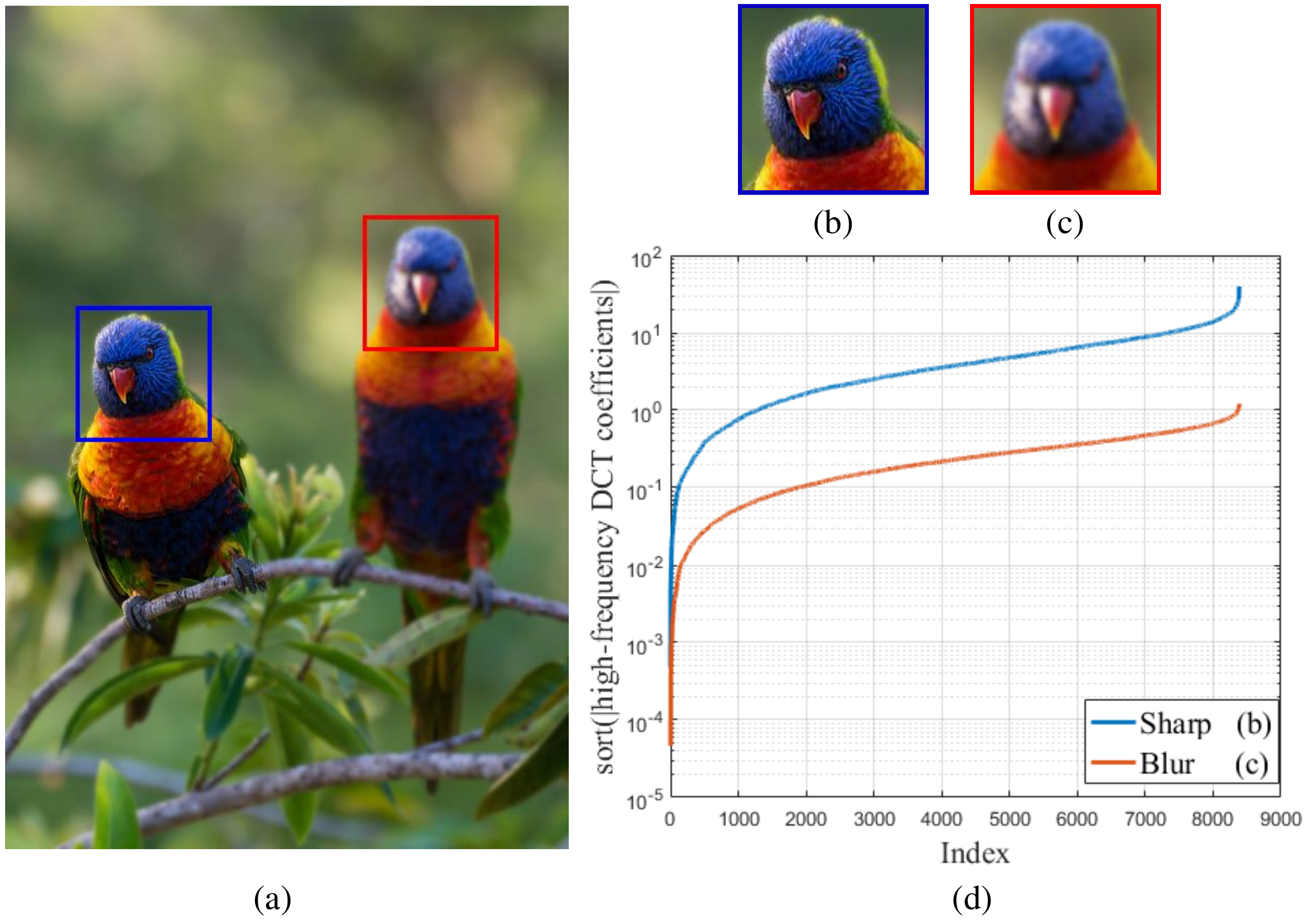}
\end{center}
   \caption{Illustration of the sorted absolute values of the high-frequency DCT coefficients  for a sharp and blurry patch in a semi-log plot. 
   (a) An input image with sharp (red patch) and blurry (blue patch) regions,
    (b) sharp patch,
     (c)~blurry patch, and
     (d) semi-log plot of the sorted absolute values of the high-frequency DCT coefficients for the sharp and blurry patches.
   }
\label{Pic2}
\end{figure}

\section{Proposed algorithm}
 To motivate our work, we first describe our proposed High-frequency  multiscale Fusion and Sort Transform (\textit{HiFST}), and then its role in image
blur detection.

\subsection{High-frequency multiscale fusion and sort transform}
The   Discrete Cosine Transform (DCT)  has emerged as one of the most popular transformations for many computer vision and image compression applications.
The DCT transforms a signal from a spatial representation into a frequency representation.
DCT coefficients can represent different frequencies, and therefore can  be informative about the image structure, energy, and bluriness.
It is well known that blur would cause a reduction in the  high frequencies of the image.
Here we divide the DCT coefficients into low-, middle-, and high-frequency bands  and consider the high-frequency  DCT coefficients.

The proposed High-frequency multiscale Fusion and Sort Transform (\textit{HiFST}) is based on computing locally at each pixel a patch-based multi-resolution (computed for different patch sizes around each pixel) DCT.
 For each pixel, the high-frequency DCT coefficients are extracted for each resolution (size of patch surrounding the considered pixel). 
 The high-frequency coefficients from all resolutions are then combined together in a vector and sorted in the order of increasing absolute values. 
 In this way, a vector of multiscale-fused and sorted high-frequency transform coefficients is generated for each image pixel.
 
Figure \ref{Pic2} shows the semi-log plot of a sharp and blurry patch to illustrate the effectiveness of the sorted absolute values of the high-frequency DCT coefficients in differentiating between a sharp and blurry region.
As shown in Figure \ref{Pic2}, after sorting the absolute values of the high-frequency  coefficients in increasing order,  obtained using the grayscale versions of  patches (a) and (b),  there is a clear visual difference between the  sorted coefficients of the blurred and unblurred patches.
In other words, as shown in Figure \ref{Pic2}, we can see that the values of the sorted coefficients in the blurry block (red block) are correspondingly smaller than the sorted coefficients values in the sharp block (blue block).
 We aim to model this property   for blur detection purposes.

Let $I$ denote the $N_1 \times N_2$-pixel input image.
We first compute the DCT of the input image in a patch-wise manner.
Let $P_{i,j}^M(i',j')$ denote a patch of size $M\times M$ centered at pixel 
$(i,j)$, with 
$i-\left\lfloor \frac{M}{2}\right\rfloor \leq i'\leq i+\left\lfloor \frac{M}{2}\right\rfloor $, and $j-\left\lfloor \frac{M}{2}\right\rfloor \leq j'\leq j+\left\lfloor \frac{M}{2}\right\rfloor 
$, where $\lfloor\frac{M}{2}\rfloor$ denotes floor of $\frac{M}{2}$.
Let $\hat{P}_{i,j}^M(\upsilon,\nu)$, $0\leq\upsilon,\nu\leq M-1
$, denote the DCT of $P_{i,j}^M(i',j')$.

\begin{figure}[t]
\begin{center}
   \includegraphics[width=0.55\linewidth]{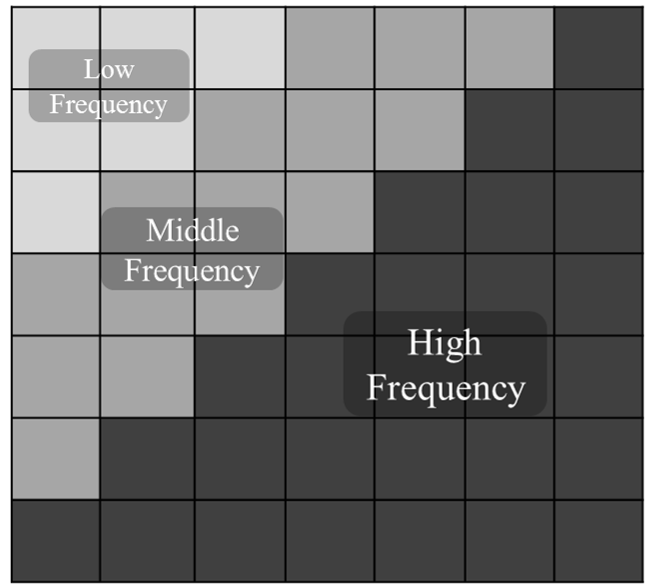}
\end{center}
   \caption{Illustration of  DCT coefficients for a $7 \times 7$ block while dividing them into  low-, middle-, and high-frequency bands.
   }
\label{Pica2}
\end{figure}

In our proposed method, we divide the computed DCT coefficients for each patch into three frequency bands, namely low, middle, and high-frequency bands \cite{langelaar2000watermarking}, and consider the high-frequency components.
Figure \ref{Pica2} illustrates the three defined frequency bands for a $7\times 7$ block.

Let $H^M_{i,j}$ denote a vector consisting of the absolute values of   the  high-frequency DCT coefficients of  $\hat{P}_{i,j}^M$. $H_{i,j}^M$ is given by: 
\begin{equation}
H_{i,j}^M=\{|\hat{P}_{i,j}^{M}(\upsilon,\nu)|:\,\upsilon+\nu\geq M-1,0\leq\upsilon,\nu\leq {M}-1\}.
\label{E1}
\end{equation}

We define the increasingly sorted vector of the absolute values of high-frequency DCT     coefficients as: 
\begin{equation}
L_{i,j}=sort(H_{i,j}^M),
\label{E2}
\end{equation}
where $L_{i,j}$ is a $1\times \frac{M^{2}+M}{2}$ vector.
Let $L_{i,j;t}$ be the $t^{th}$ element in vector $L_{i,j}$ and let $L_t$ be the $t^{th}$ layer that is obtained by grouping all the $t^{th}$ elements $L_{i,j;t}$ for all positions $(i,j)$. $L_t$ can be represented as an $N_1\times N_2$ matrix given by:
\begin{equation}
L_t= \{{L_{i,j;t}}, 0\leq i<N_{1}, 0\leq j<N_{2}\}.
\label{E3}
\end{equation}

The proposed \textit{HiFST} consists of the $\frac{M^{2}+M}{2}$ normalized layers $\hat{L}_t, 1\leq t \leq\frac{M^{2}+M}{2}$, where $\hat{L}_t$ is given by:
\begin{equation}
\hat{L}_{t}=\frac{L_{t}-\min(L_{t})}{\max(L_{t})-\min(L_{t})},\;1\leq  t \leq \frac{M^{2}+M}{2}.
\label{E4}
\end{equation}

By normalizing each layer between $[0,1]$, each layer can better differentiate between the blurred and unblurred regions in the image  and measure locally the level of blur. 
 In Eqs.~(\ref{E2})-(\ref{E3}), we differentiate between the blurred and unblurred regions in a local adaptive manner by extracting the  high-frequency DCT coefficients in a block-wise manner and then grouping and normalizing sorted DCT 
coefficients belonging to the same position.

\begin{figure}[t]
\begin{center}
   \includegraphics[width=1.0\linewidth]{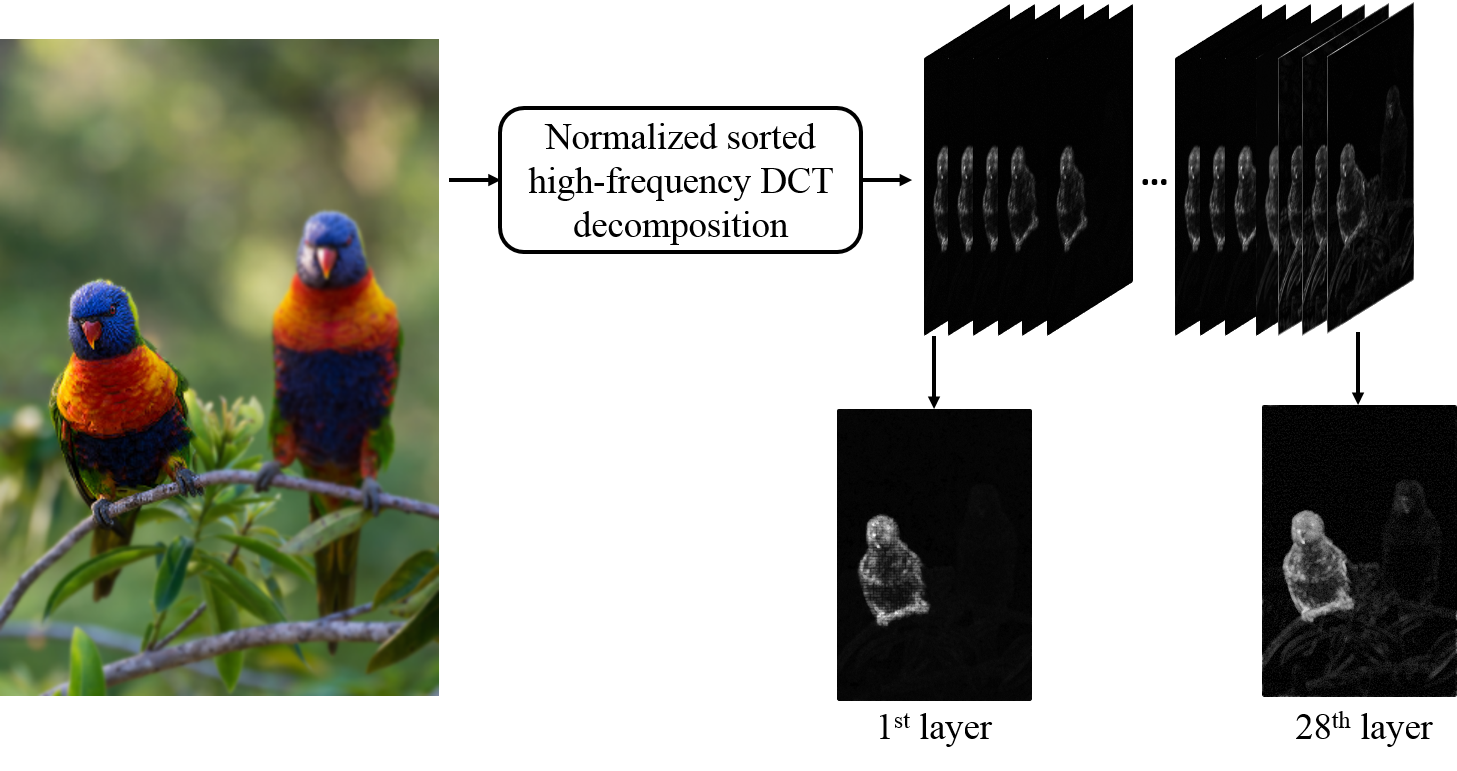}
\end{center}
   \caption{Illustration  of the normalized-sorted high-frequency decomposition layers for a blurry image, where the unblurred regions have larger values than the blurred ones.}
\label{Pic3}
\end{figure}

In Figure \ref{Pic3} for illustration purposes, we employ a normalized sorted high-frequency DCT decomposition on each $7\times 7$ block with one block located at each pixel, which leads to   28  layers.
The first  layer contains the normalized smallest high frequency values, which yields to differentiating between the sharp and blurry regions.
As we move toward the higher layers which consist of       larger high-frequency coefficients, we can see more structures and edges appear for  both the sharp and blurred regions   while the discrimination between blurred and unblurred regions is  still noticeable.

Considering  only one resolution  may not  accurately indicate whether an image or patch is blurred or not. 
The scale ambiguity has been studied in various applications \cite{shi2014discriminative,yan2013hierarchical}. 
Therefore, here we take into account a multiscale model to fuse information from different scales.
With our multiscale analysis, our method is able to deal with blur in either small-scale or large-scale structures, so that blurred and unblurred regions are detected more effectively.
We utilize a multiscale  \textit{HiFST}  decomposition as follows:
\begin{equation}
L_{i,j}=sort(\bigcup_{r=1}^{m}{H}_{i,j}^{M_{r}}),
\label{E7}
\end{equation}
where   $M_r=2^{2+r}$ if even and $M_r=2^{2+r}-1$ if odd,  $\bigcup_{r=1}^{m}$ denotes the union of all the high-frequency DCT coefficients computed in $m$ different scales with different resolutions, and   $L_{i,j}$ is a $1 \times{\sum_{r=1}^{m}\frac{{M_r}^2+M_r}{2}}$ vector.
Then   ${L_t}$ and $\hat{L}_{t}$  can be computed as described in  Eqs. (\ref{E3}) and (\ref{E4}).

 \subsection{Proposed spatially-varying blur  detection } 
In the following, we present in details our spatially-varying blur detection approach which is based on the fusion, sorting, and normalization of multiscale high-frequency
DCT coefficients 
 of gradient magnitudes  to detect blurred and unblurred regions from a single image without having any information about the camera settings or the blur type.
The flowchart of our proposed algorithm is provided in Figure \ref{Pic4}. 
A blurred image, $B$, can be modeled as follows:
\begin{equation}
B=A\ast b+n,
\end{equation}
where $A$ is a sharp latent image, $b$ is the blur kernel, $\ast$  is the convolution
operator, and $n$ is the camera noise. Our goal is to estimate the blur map from the observed blurry image $B$.
Given the  image $B$, we first apply a Gaussian filter with a small kernel to   remove the high-frequency noise.
The Gaussian filter is given by:

\begin{equation}
g(i,j;\sigma)=\frac{1}{{2\pi\sigma^{2}}}\exp(-\frac{i^{2}+j^{2}}{2\sigma^{2}}),
\end{equation}
where $\sigma$ is set to 0.5 in our experiment. 
 Let $B_g$ denote the Gaussian filtered image of the input blurry image $B$.
The gradient magnitudes of an image can effectively capture image local structures, to which the human visual system is highly sensitive.
By computing the gradient magnitudes, most of the spatial redundancy is removed, and the image structure and shape components are preserved. 
The gradient magnitude image $G$ of the Gaussian filtered image,  $B_g$, is  computed by:

  \begin{figure}[t]
\begin{center}
   \includegraphics[width=0.98 \linewidth]{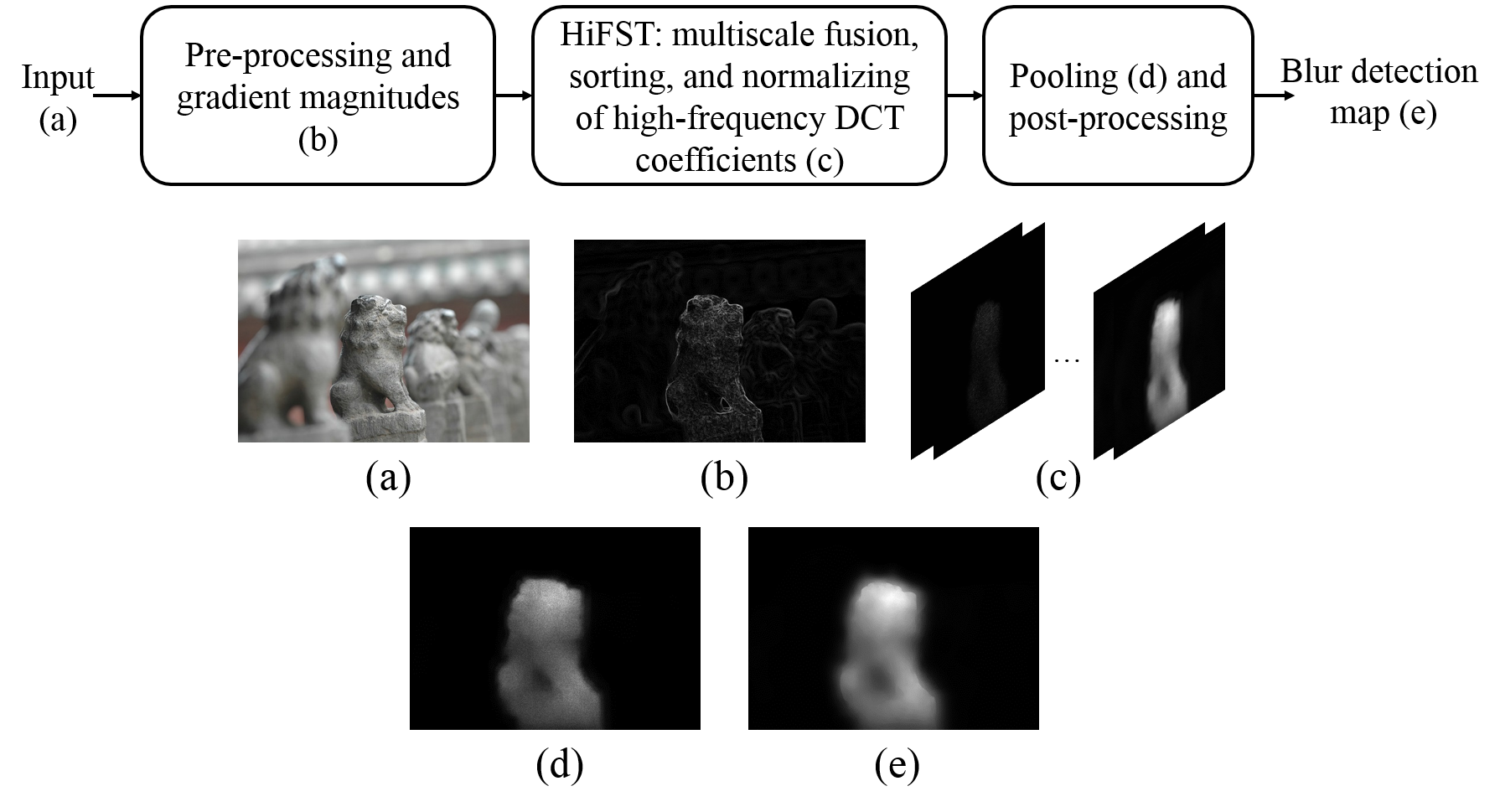}
\end{center}
   \caption{Flowchart of our proposed blur detection algorithm along with example outputs for each stage.}
\label{Pic4}
\end{figure}

\begin{equation}
G=\sqrt{(B_{g}\ast h_{x})^{2}+(B_{g}\ast h_{y})^{2}},
\end{equation}
where $h_{x}=\left[\begin{array}{lr}
1 & 0\\
0 & -1
\end{array}\right]$, and $h_{y}=\left[\begin{array}{rr}
0 & 1\\
-1 & 0
\end{array}\right]$.

\begin{figure}[]
\begin{center}
   \includegraphics[width=0.98 \linewidth]{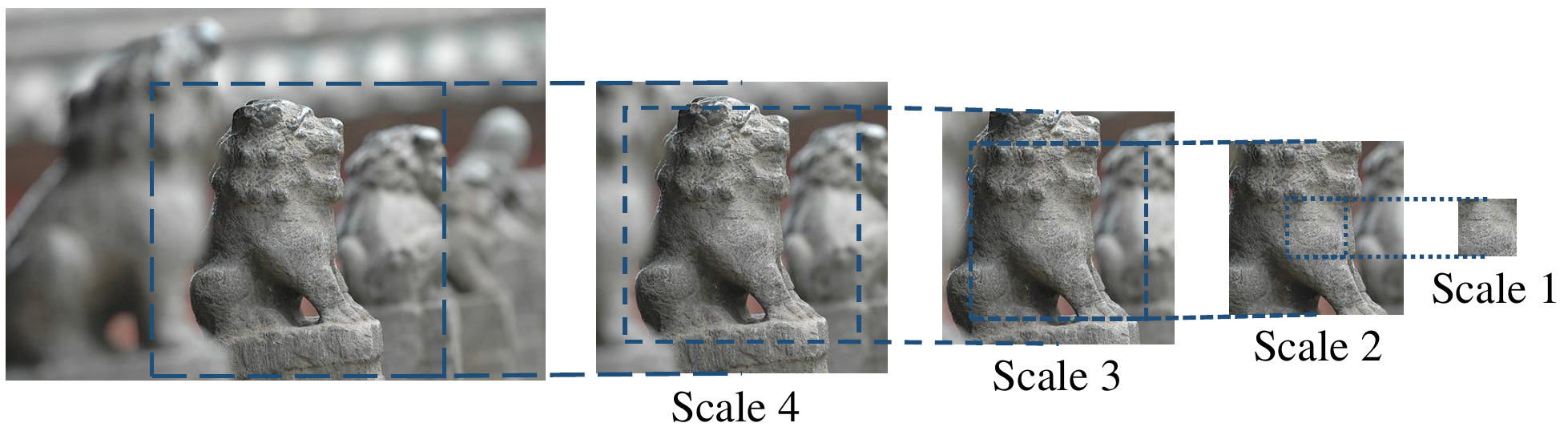}
\end{center}
   \caption{Demonstration of multiscale blur perception for different scales. 
The blur confidence   highly depends on the patch scale.}
\label{Pic5}
\end{figure}

Next we apply the \textit{HiFST} (Section 2.1) decomposition
on the computed gradient image, $G$, in a multiscale manner.
 As shown in Figure \ref{Pic5}, blur can be perceived differently in different scales.
Given an image $G$, we compute its multiscale 
\textit{HiFST} as described in Section 2.1 where the image $I$ is 
now replaced with the gradient magnitude image $G$.
We compute  ${L_t}$ and  $\hat{L}_{t}$ as described in Section 2.1, where   $m=4$.
To compute our proposed blur detection map,
 we consider  only the  first $\sum_{r=1}^{m}M_{r}$ layers. 
 Based on our experiment and observation, using the first $\sum_{r=1}^{m}M_{r}$ layers\textsuperscript{\ref{FF1}} of the \textit{HiFST} provides sufficient information to compute the blur map.  
Our proposed blur detection map, $D$, is computed as follows:

\begin{subequations}
 \begin{equation}
D=T\circ\:\omega,
\end{equation}   
where $\circ$ denotes the Hadamard product of matrices $T$ and $w$ whose  elements are given by:
\begin{equation}
  \begin{multlined}
T_{i,j}=\max(\{\hat{L}_{i,j;t}:\;t=1,...,\sum_{r=1}^{m}M_{r}\}),\\
 0\leq i<N_{1}, 0\leq j<N_{2},
  \end{multlined}
\end{equation}    
\begin{equation}
\omega_{i,j}=-\sum_{(i',j')\in R_{(i,j)}^{k}}P(T_{i',j'})\log[{P}(T_{i',j'})],
\end{equation}    
 \end{subequations}
where $R_{(i,j)}^k$ denotes a $k\times k$ patch centered at
pixel $(i,j)$, and $P$  denotes a probability function.
From Eqs. (9b) and (9c), it can be seen that   $T$ is obtained by max pooling, and  $\omega$ is obtained by computing the  local entropy map  of $T$, using a  $k\times k$ neighborhood around the corresponding pixel in   $T$ 
(in our implementation, we use $k=7$)\footnote{
The selection
of this value is not critical. The results are very close
when the value is chosen within a  $\pm20\%$ range.\label{FF1}}.
The entropy map ($\omega$) is used as  a weighting factor to  give more weight to the salient regions in the image.
The final blur map is smoothed using edge-preserving filters \cite{gastal2011domain} to suppress the influence of outliers  and preserve boundaries. 

We further extend our proposed algorithm to estimate the camera focus points map.
The camera focus points map shows the focus points of the camera while taking a photo.
Intuitively, this region should have the highest intensity  in the  blur detection map.
We compute the camera focus points map  by  as follows:

\begin{equation}
F=\begin{cases}
\begin{array}{c}
1\\
0
\end{array} & \begin{array}{c}
D'\geq Th\\
D'<Th
\end{array}\end{cases},
\end{equation}    
where $F$ denotes the  camera focus points map and $D'$ is a Gaussian smoothed version of $D$, normalized between [0,1]. 
 In our experiment, we set the threshod value, $Th$, to 0.98.\textsuperscript{\ref{FF1}}

\section{Results}
In this section we evaluate the performance of our proposed method, \textit{HiFST}, quantitatively and qualitatively.
Here we show that our proposed method outperforms the existing algorithms in terms of both quantitative and qualitative results regardless of the blur type. 
Moreover, we  evaluate the robustness of our method to  different types and levels of distortions.

 \begin{figure}[t]
\begin{center}
      \includegraphics[width=0.98\linewidth]{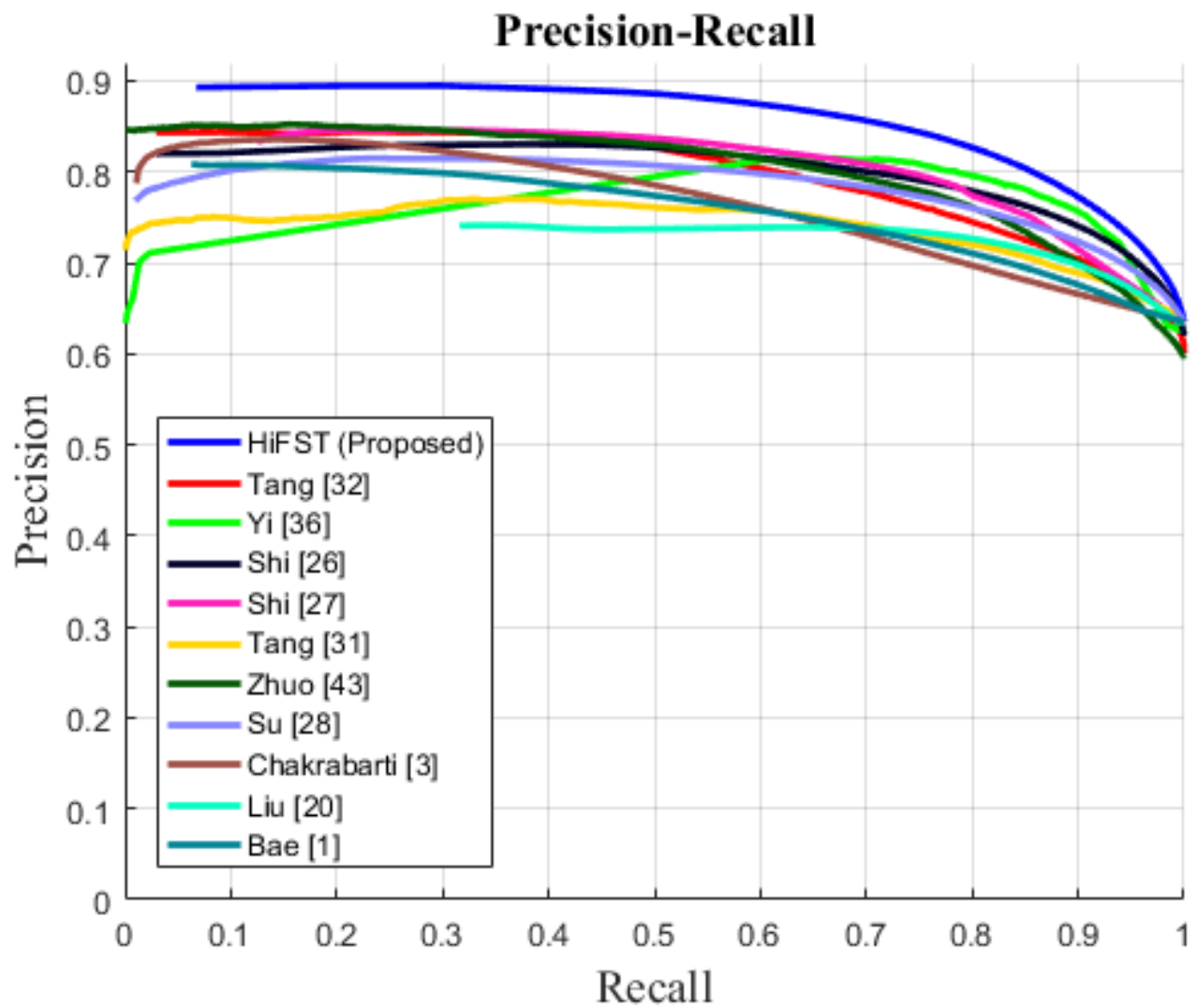}
\end{center}
   \caption{Quantitative  Precision-Recall comparison on the blur  dataset \cite{shi2014discriminative} for
different methods.}
\label{Pic6}
\end{figure}

\begin{figure*}[t!]
\begin{center}
\includegraphics[width=0.99 \linewidth]{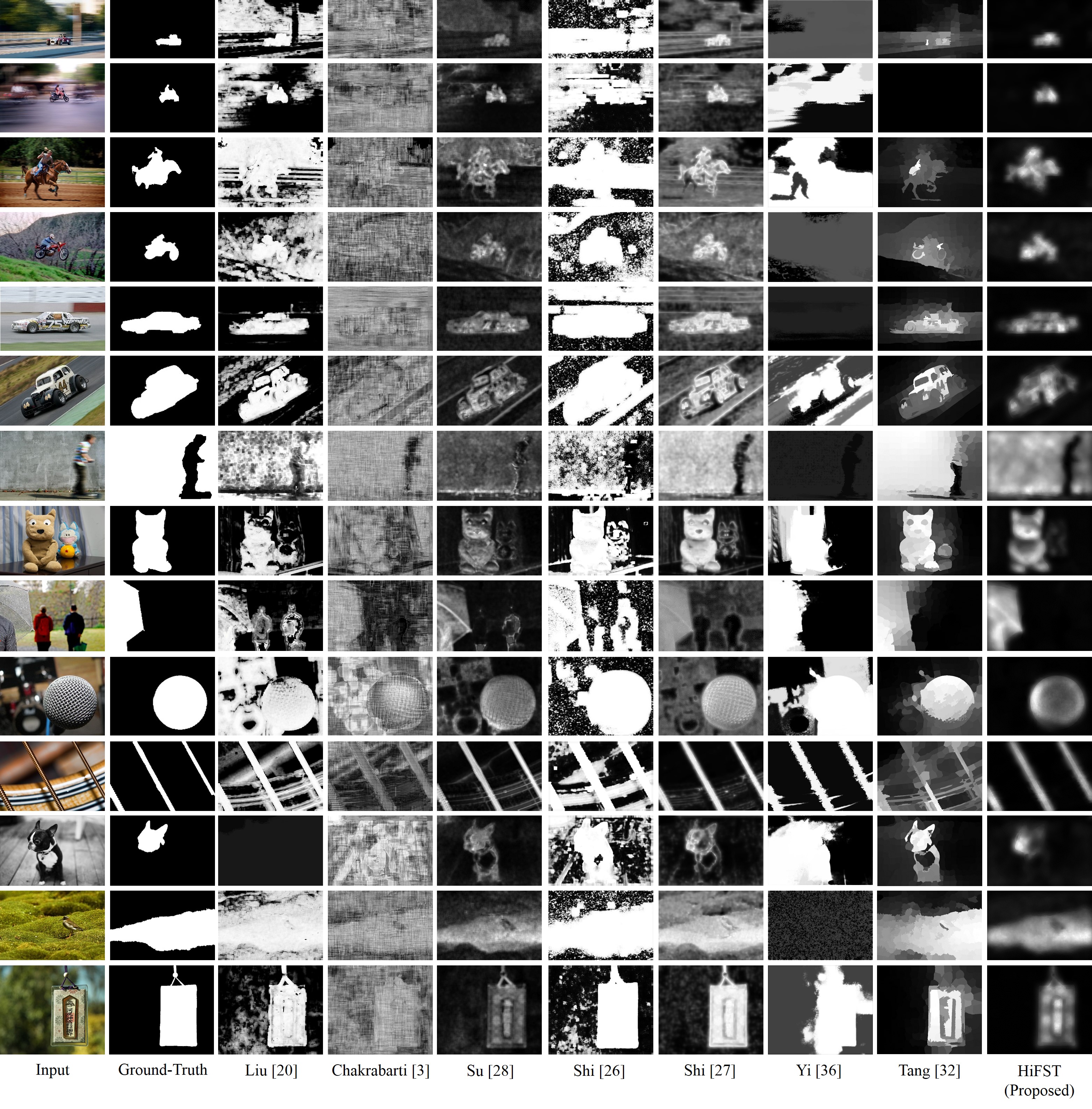} 
\end{center}
   \caption{Visual comparison on images selected from \cite{shi2014discriminative,shi2015just} while adding zero-mean Gaussian noise with $\sigma^2=10^{-4}$ to the input images.}
   \label{Pic8}
\end{figure*}

 For quantitative comparison, we evaluate our method on a variety of images with different types and amount of blur provided in dataset \cite{shi2014discriminative} and compare our results with state-of-the-art algorithms  \cite{bae2007defocus,chakrabarti2010analyzing,liu2008image,shi2014discriminative,
shi2015just,su2011blurred,tang2013defocus,tang2016spectral,
 yi2016lbp,zhuo2011defocus}. 
 Figure \ref{Pic6} shows the precision-recall curve for the blur dataset \cite{shi2014discriminative}, which consists of 296 images with motion blur and 704 images with defocus blur.
In our experiment, we binarized the  blur detection map by varying the threshold    within the range [0, 255].
Our proposed method achieves the highest precision within  the entire recall
range [0, 1], which conveys its potential for different levels and types of blur.

In Figure \ref{Pic8}, we evaluate the performance of our method qualitatively on images provided in \cite{shi2014discriminative,shi2015just} with different contents as well as different types and levels of blur, in which we added zero-mean Gaussian noise with $\sigma^2=10^{-4}$ to take into account the camera noise.
Although the amount of noise is not easily noticeable  by human eyes, it  simulates the practical camera noise case.
We evaluate the performance of the proposed algorithm against the state-of-the-art methods \cite{chakrabarti2010analyzing, liu2008image,shi2014discriminative, shi2015just, su2011blurred,tang2016spectral,yi2016lbp}\footnote{More visual results as well as our MATLAB code  are available at: \url{http://ivulab.asu.edu/software}.}. In the provided maps,  the unblurred regions have higher intensities than the blurred ones.
As shown in Figure \ref{Pic8}, our proposed method can handle different types of blur and can distinguish between the blurred and unblurred regions effectively.
Algorithms \cite{chakrabarti2010analyzing,  yi2016lbp} dramatically failed due  to their sensitivity to the noise or type of blur.
Moreover,  although algorithms \cite{liu2008image, shi2014discriminative,shi2015just, su2011blurred,tang2016spectral} can  detect the blur map for some of the images, their detected maps include incorrectly labeled regions compared to the ground-truth.
 In contrast, our proposed  method   can distinguish between the blurred and unblurred regions with high accuracy regardless of the blur type.

  \begin{figure}[t]
\begin{center}
      \includegraphics[width=0.95 \linewidth]{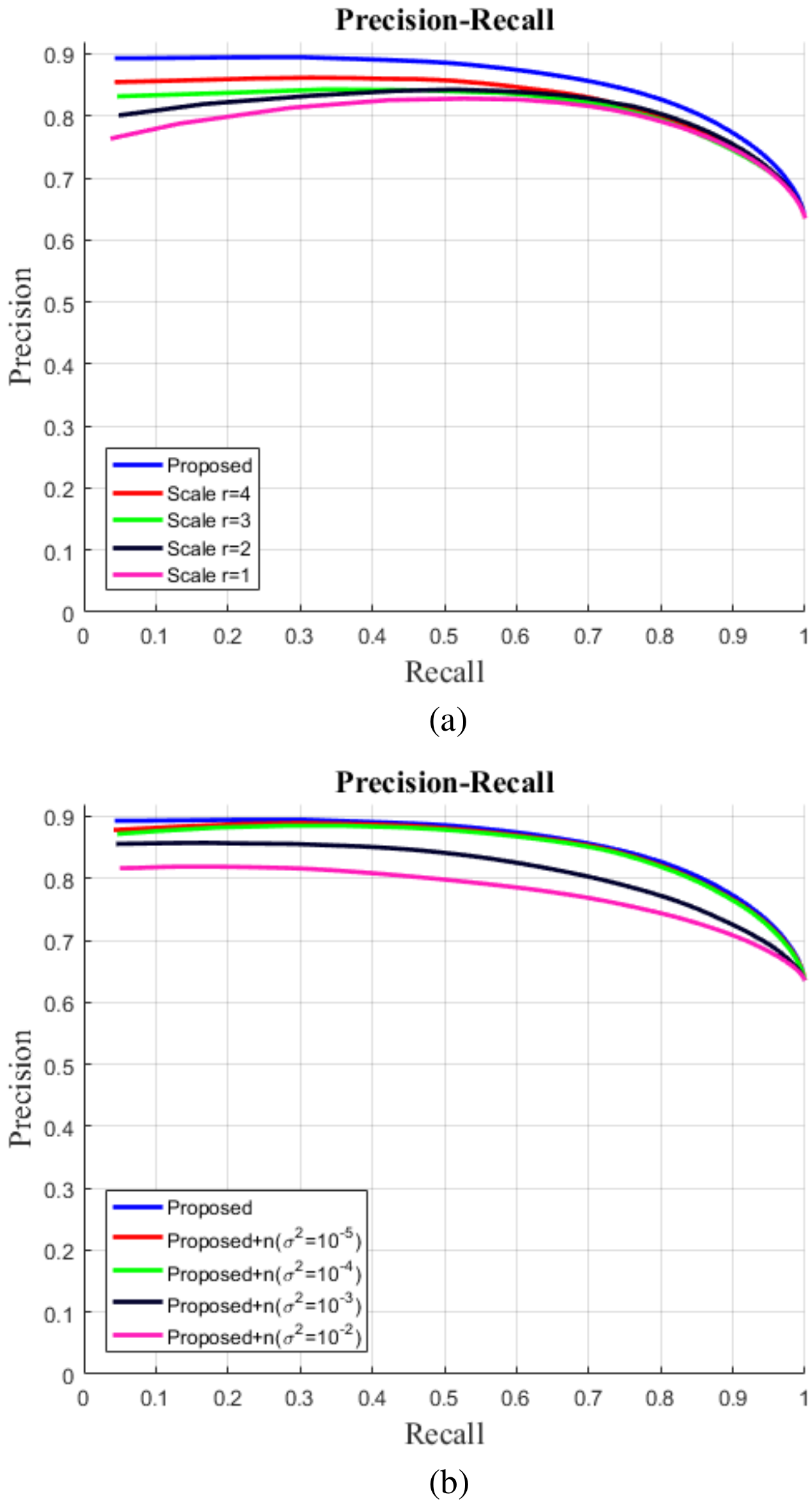}
\end{center}
   \caption{(a) Precision-Recall curves between our proposed method in a multiscale manner and the results of our method via each scale separately.
 (b) Precision-Recall comparison of our proposed method on the blur  dataset \cite{shi2014discriminative} in which  we add zero-mean Gaussian noise with different variance $\sigma^2=\{0, 10^{-5},  10^{-4},  10^{-3},  10^{-2}\}$ to the   input image.}
\label{Pic66}
\end{figure}
Furthermore, Figure \ref{Pic66} (a) demonstrates the effectiveness of our multiscale approach by comparing it to just using one single scale.
We evaluate the precision-recall curves resulting from our method when it just uses one scale at a  time ($M_r=2^{2+r}-1, 1\leq r \leq4$)  and compare it to our final multiscale method. 
As shown in Figure \ref{Pic66} (a)  employing all the scales  leads to the best results.
Furthermore, to validate the robustness of our method qualitatively and quantitatively, 
we test our method on dataset \cite{shi2014discriminative} while adding zero-mean Gaussian noise with different densities to the input images.
In our experiment the variance ($\sigma^2$) of the noise is varied between   zero to $10^{-2}$.
As shown in Figure \ref{Pic66}(b), the resulting precision-recall curves in the presence of noise with  ${\sigma^2=\{ 0, 10^{-5}, 10^{-4}\}}$  are almost the same.
By adding noise with   larger variances, ${\sigma^2=10^{-3}}$ and ${\sigma^2=10^{-2}}$, the precision-recall curves show only a slight drop in performance as compared to the noise-free case and show that 
the proposed method still achieves competitive results.

\begin{figure}[h]
\begin{center}
 \includegraphics[width=1\linewidth]{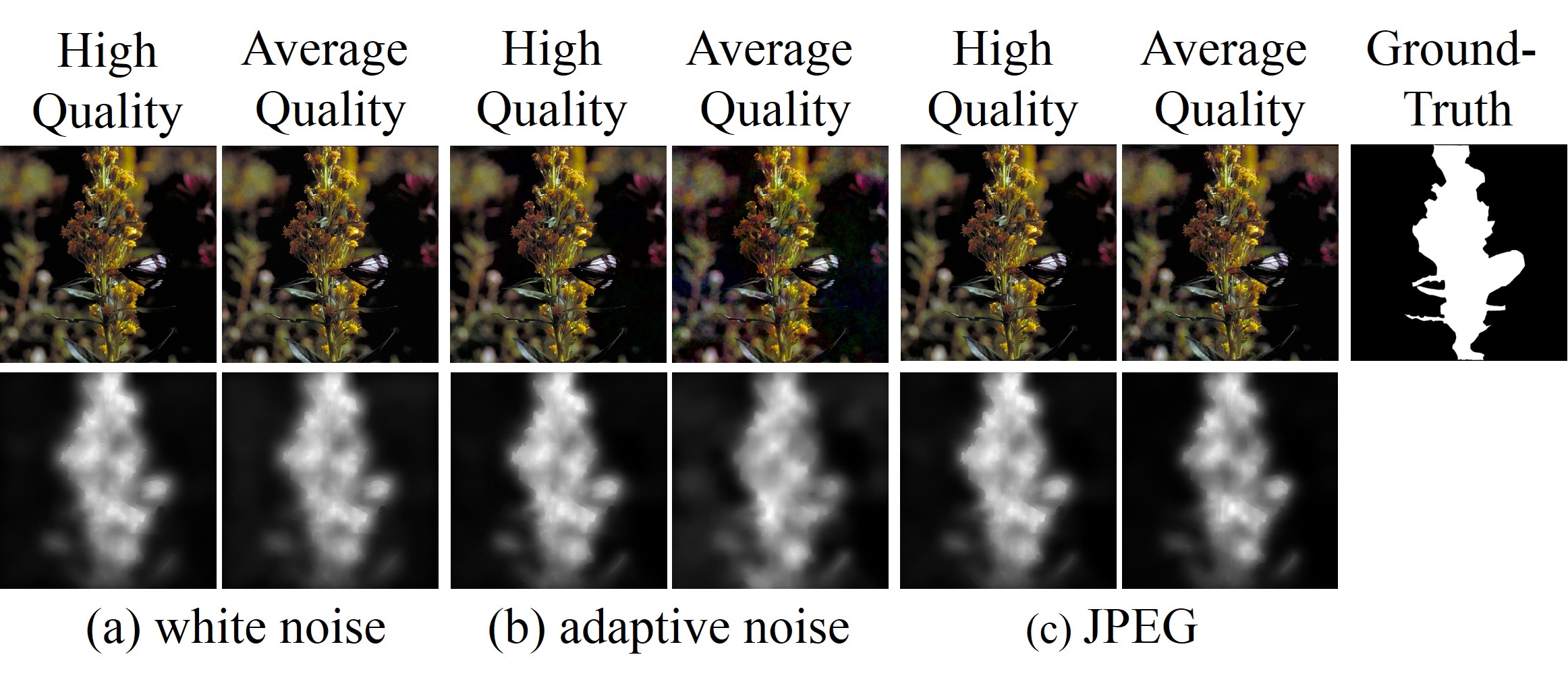}
\end{center}
\vspace{-2mm}
   \caption{Qualitative  evaluation of the robustness of our proposed method to different types and levels of distortions for images in the CSIQ database \cite{larson2010most}.
   (a) Distorted images under different levels of white noise.
   (b) Distorted images under different levels of adaptive noise.
   (c) Distorted images under different levels of JPEG distortion.
}
         \label{Pic11}
\end{figure}

In Figure \ref{Pic11}, we evaluate the robustness of our method qualitatively on images from the Categorical Image Quality (CSIQ) database\cite{larson2010most} with  
 different types and levels of distortions,  such as zero-mean Gaussian noise, adaptive noise, and JPEG.
 As shown in Figure \ref{Pic11},  our proposed method has the capability of estimating  the blur map for distorted images.

\begin{figure}[h]
\begin{center}
   \includegraphics[width=1 \linewidth]{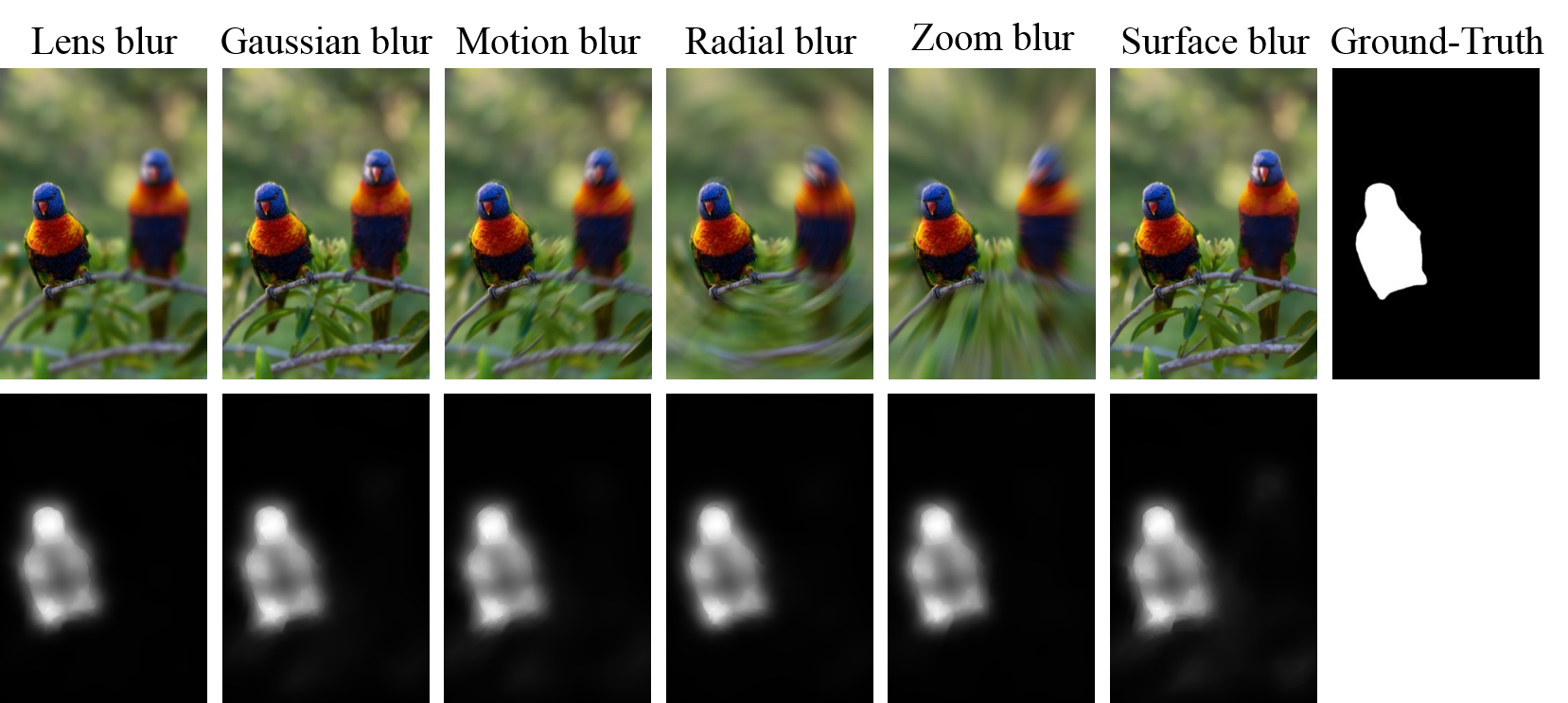}
\end{center}
   \caption{Visual  results of our proposed method on images with different types of blur.}
      \label{Pic9}
\end{figure}

Finally, to further demonstrate the performance of our proposed method for different blur types, we test our algorithms on 6 synthetic examples of an image with different types of blur on the background region, such as lens, Gaussian, motion, radial, zoom, and surface blur.
 As shown in Figure  \ref{Pic9}, our proposed method can handle all these blur types    accurately compared to the ground-truth.

\section{Applications}
In this section, we demonstrate the use of  our proposed blur detection map in a few applications, such as blur magnification, deblurring, depth of  field (DOF), depth from focus (DFF), and camera focus points estimation.

\subsection{Blur Magnification} 
\begin{figure}[h]
\begin{center}
   \includegraphics[width=0.96\linewidth]{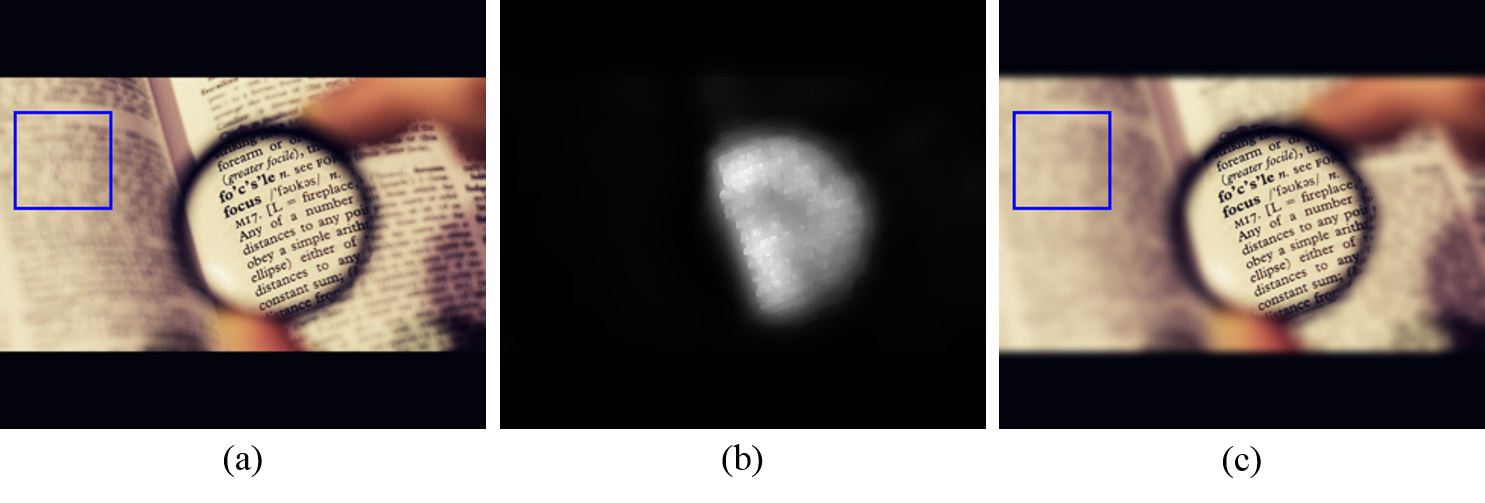}
\end{center}
   \caption{Blur magnification. (a) Input image. (b) Our estimated blur map. (c) Results after blur magnification using our proposed blur map.}
      \label{Pic10}
\end{figure}

Given the estimated blur map, we
can perform blur magnification \cite{bae2007defocus}.
 Blur magnification increases the level of blur in out-of-focus areas and makes the image have a shallower DOF.
 Therefore, the in-focus object would be highlighted more. 
 Figure \ref{Pic10} shows   examples of blur magnification by using our detected blur map. 

\subsection{Deblurring}
\begin{figure}[h]
\begin{center}
   \includegraphics[width=0.99\linewidth]{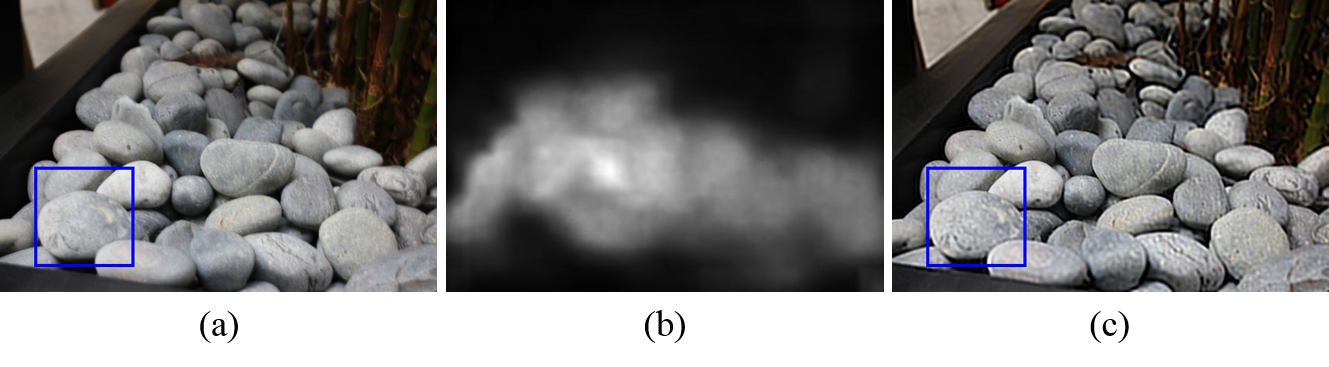}
\end{center}
   \caption{Application of our proposed method in deblurring.
    (a)~Input image. (b) Our estimated blur map. (c) Deblurred results.}
            \label{Pic14}
\end{figure}

In Figure \ref{Pic14}, we use our estimated blur map (Figure \ref{Pic14}(b)) in the deblurring algorithm described in \cite{shen2012spatially} and recover the clear image; Figure \ref{Pic14}(c) represents the deblurring result.

\subsection{Depth of field estimation}
\begin{figure}[]
\begin{center}
   \includegraphics[width=0.97\linewidth]{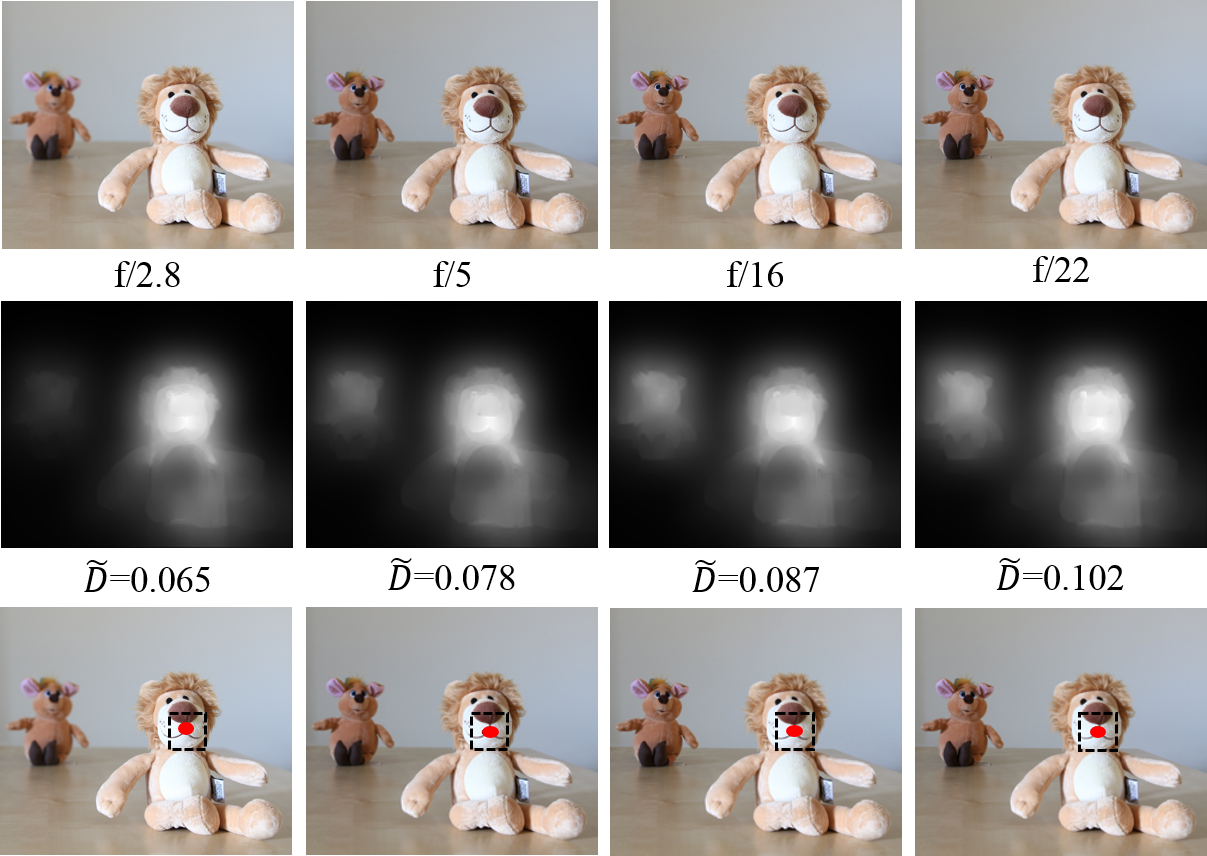}
\end{center}
   \caption{Application of our proposed method in detecting blur   for a photo   taken under different aperture sizes in a dynamic setting.
   First row: input images. Second row: detected blur maps; $\tilde{D}$ denotes the median value of the normalized
blur map $D$.  Third row: estimated camera focus points.}
            \label{Pic12}
\end{figure}

Depth of Field (DOF) refers to the area of the picture that is in-focus in an image. 
It is determined by three factors,  including   aperture size, distance from the lens, and the focal length of the lens. 
 Let  $\tilde{D}$ denote the median of the normalized blur map $D$, and be used  as an estimation for  the DOF.
In Figure \ref{Pic12}, we provide four images which are taken   with the same camera   and distance from the objects.
In our experiment, the camera focus is set to be on the face of the front object  and the  aperture size changes by choosing different f-stop, such as f/2.8, f/5,  f/16, and f/22.
 As shown in Figure \ref{Pic12}, by decreasing the aperture size (increasing the f-stop) the DOF increases and our blur map as well as $\tilde{D}$ change in a consistent way.
 Also as shown in the third row of   Figure \ref{Pic12}, the camera focus points map (red spot inside the square) for all the images stay consistent.

\subsection{Depth from focus}
\begin{figure}[h]
\begin{center}
   \includegraphics[width=0.99\linewidth]{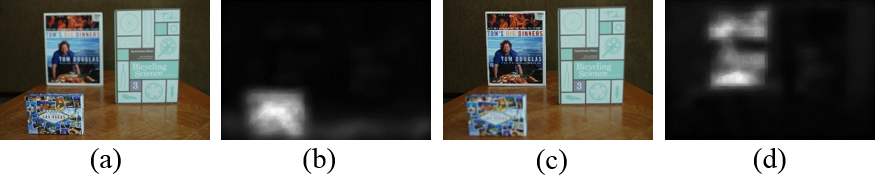}
\end{center}
   \caption{Application of our method to changing
the camera focus and DFF. (a) $\&$ (c) are images from \cite{suwajanakorn2015depth} with different focus areas. (b) $\&$ (d) are our estimated maps.}
            \label{PS_S_1}
\end{figure}

In Figure  \ref{PS_S_1}, we provide the application of our proposed method for depth
from focus (DFF).
As shown
in Figure  \ref{PS_S_1}, by changing the camera focus,   our estimated blur  maps
change in a consistent way. 

\section{Conclusion}
In this paper we have addressed the challenging problem of    blur detection from a single image without having any information about the blur type or the camera settings. 
We proposed an effective   blur detection method based on a high-frequency multiscale fusion and sort transform, which makes   use of  high-frequency DCT coefficients   of the gradient magnitudes from multiple resolutions.
Our algorithm achieves state-of-the-art results on   blurred images with different blur types and blur levels. 
To analyze the potential of our method, we also evaluated it on images with different types of blur as well as different levels and types of distortions.
Furthermore, we showed that the proposed method can   benefit different computer vision applications including camera focus points map estimation, blur magnification,  depth of field estimation, depth from focus, and deblurring.

\end{document}